\documentclass{article}
\usepackage{ijcai24}

\usepackage{times}
\usepackage{soul}
\usepackage[hidelinks]{hyperref}
\usepackage[utf8]{inputenc}
\usepackage[small]{caption}
\usepackage{graphicx}
\usepackage{amsmath}
\usepackage{amsthm}
\usepackage{booktabs}
\usepackage{algorithm}
\usepackage{algorithmic}
\usepackage[switch]{lineno}
\usepackage{xcolor}
\usepackage{xurl}
\usepackage{siunitx}

\title{From 2D to 3D: AISG-SLA Visual Localization Challenge}

\author{
Jialin Gao$^1$
\and
Bill Ong$^1$\and
Darld Lwi$^{2}$\and
Zhen Hao Ng$^2$\and
Xun Wei Yee$^1$\and
Mun-Thye Mak$^1$\and
Wee Siong Ng$^3$\and
See-Kiong Ng$^4$\and
Hui Ying Teo$^2$\and
Victor Khoo$^2$\and
Georg Bökman$^5$\and
Johan Edstedt$^6$\and
Kirill Brodt$^7$\and
Clémentin Boittiaux$^8$\and
Maxime Ferrera$^8$\And
Stepan Konev$^9$\\
\affiliations
$^1$AI Singapore, Singapore\\
$^2$Singapore Land Authority, Singapore\\
$^3$Institute for Infocomm Research, Singapore\\
$^4$National University of Singapore, Singapore\\
$^5$Chalmers University of Technology, Sweden\\
$^6$Link\"{o}ping University, Sweden\\
$^7$Universit\'{e} de Montr\'{e}al, Canada\\
$^8$Ifremer, Centre M\'{e}diterran\'{e}e, France\\
$^9$Booking.com, Netherlands\\
\emails
\{jialin, bill\_ong\}@aisingapore.org, 
\{darld\_lwi, ng\_zhen\_hao\}@sla.gov.sg, 
\{xunwei, munthye\}@aisingapore.org, 
wsng@i2r.a-star.edu.sg, 
seekiong@nus.edu.sg, 
\{teo\_hui\_ying, vic-tor\_khoo\}@sla.gov.sg, 
bokman@chalmers.se, johan.edstedt@liu.se, kirill.brodt@umontreal.ca, 
\{boittiauxclementin, maxime.ferrera, stevenkonev\}@gmail.com
}

\begin{document}

\maketitle

\begin{abstract}
    Research in 3D mapping is crucial for smart city applications, yet the cost of acquiring 3D data often hinders progress. Visual localization, particularly monocular camera position estimation, offers a solution by determining the camera's pose solely through visual cues. However, this task is challenging due to limited data from a single camera. To tackle these challenges, we organized the AISG–SLA Visual Localization Challenge (VLC) at IJCAI 2023 to explore how AI can accurately extract camera pose data from 2D images in 3D space. The challenge attracted over 300 participants worldwide, forming 50+ teams. Winning teams achieved high accuracy in pose estimation using images from a car-mounted camera with low frame rates. The VLC dataset is available for research purposes upon request via {\it vlc-dataset@aisingapore.org}.
\end{abstract}

\section{Introduction}

Camera pose estimation, also referred to as visual localization~\cite{barros2022comprehensive}, plays a pivotal role in determining the 6-degree-of-freedom pose (3D position and orientation) of a camera within its environment using visual cues. Specifically, monocular camera poses estimation~\cite{engel2014lsd} involves utilizing a single camera to derive the camera's position and orientation. This process entails analyzing 2D images captured by the camera and computing a transformation matrix that characterizes the camera's spatial relationship to the 3D world. Despite the challenges posed by the limited information available from a single camera, the significance of camera pose estimation cannot be overstated, especially considering the high costs associated with obtaining 3D data from alternative sources like LiDAR sensors.

Accurate camera pose estimation is indispensable across a spectrum of real-world applications, ranging from urban planning~\cite{coors2000matching} to augmented reality, underwater surveillance~\cite{gonzalez2023survey}, robotics, and autonomous vehicles. In urban planning, camera pose estimation equips planners with valuable spatial insights that inform decision-making processes. In augmented reality, precise knowledge of the user's device position and orientation relative to the surrounding environment is pivotal for convincingly overlaying virtual objects. Moreover, camera pose estimation plays a critical role in underwater surveillance by facilitating the tracking and monitoring of objects, structures, and environmental conditions in aquatic environments. Furthermore, in robotics and autonomous vehicles, accurate camera pose estimation empowers robots to navigate and interact with their surroundings effectively.

To promote the development and testing of models in more dynamic and diverse environments, we introduce the AISG–SLA Visual Localization Challenge (VLC). The primary objective of the challenge is to discern the relative pose estimates among images obtained by a monocular camera affixed to a vehicle navigating the streets of Singapore. The focus lies in accurately measuring the rotational differences between successive images, reflecting the subtle variations in orientation as the vehicle progresses along its path. In addition, the secondary objective involves precisely estimating relative translations, and capturing the spatial movements and displacements between scenes observed in consecutive frames. These two assessment criteria require participants to address the complexities of visual localization in dynamic urban settings effectively for both rotational alignment and positional tracking within real-world driving scenarios. 

The paper's structure is as follows: Section 2 reviews existing camera pose estimation studies. Section 3 explores the VLC dataset's characteristics and challenges. Section 4 outlines winning teams' strategies~\footnote{\url{https://prizechallenge.aisingapore.org/competitions/1/visual-localisation/leaderboard/}}. The conclusion offers final remarks.
\section{Related Work}

Within the realm of camera pose estimation, there exist two primary approaches for solving the associated problems: direct~\cite{engel2017direct} and indirect~\cite{mur2015orb} methods. Indirect methods typically involve detecting interest points, associating them with feature descriptors, and optimizing the camera pose and 3D point clouds by minimizing reprojection error~\cite{rosinol2020kimera,campos2021orb}. On the other hand, direct methods delve into the image formation process and define objectives based on photometric error~\cite{zubizarreta2020direct}. While direct methods capture more image details, such as lines and intensity variations~\cite{engel2017direct}, they face more complex optimization challenges and are less robust to geometric distortions.

Recent advancements in deep learning-based techniques~\cite{zhou2018deeptam} have shown promise in addressing challenging scenarios in camera pose estimation. These techniques often involve training systems for specific subtasks, such as feature detection, matching, and localization. However, some deep learning models tend to concentrate on small-scale reconstruction and may lack capabilities like loop closure and global bundle adjustment, limiting their applicability for large-scale deployment.

Current methodologies heavily rely on training deep models using datasets such as KITTI~\cite{geiger2013vision}, EuRoC~\cite{burri2016euroc}, and TartanAir~\cite{wang2020tartanair}. However, these datasets often exhibit limited motion amplitude and may contain synthetic data, which may not accurately represent real-world application scenarios.

\section{VLC Dataset}
In this section, we outline the characteristics of the dataset. 

\subsection{Data Sources}

The dataset utilized in this research comprises photographs taken by a spherical camera system equipped with a 5-megapixel resolution and a Sony IMX264 $2/3''$ CMOS sensor, featuring a pixel size of $3.45 \times 10-3$ mm and utilizing 4.4 mm focal length lenses. These photographs were captured within the urban areas of two different townships in Singapore, primarily showcasing human subjects (faces captured in images were blurred), vehicles, architectural structures, and natural landscapes predominantly found within residential areas and parks. 

The images were captured based on distance (i.e., every 10 meters), estimated by the number of tire rotations. This is approximately equivalent to 1 or 2 shots every second. Some distances between timestamps might have a larger difference (e.g., 8.8 meters vs. 10 meters) because the car might have been making a turn. The visual data for the VLC was provided in the form of street-level monocular images stored in JPEG format. The dataset was divided into two sets: a training set containing 10,007 images and a testing set containing 2,219 images. These sets were further subdivided into trajectories, with four trajectories allocated for training purposes and one trajectory reserved for testing. Here, a trajectory refers to a sequence of images captured as the camera moves along a continuous path. The intrinsic parameters, including the focal length and optical center (also known as the principal point), were provided.

\subsection{Data Matching Challenge}

\begin{figure}
    \centering
    \includegraphics[width=\linewidth]{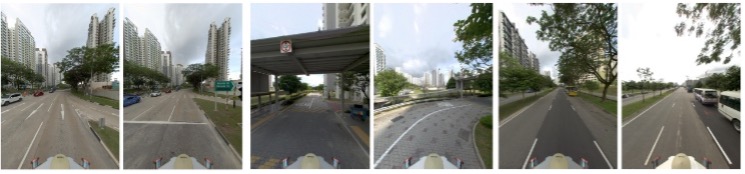}
    \caption{The camera is directed towards the rear of the vehicle, resulting in consecutive captures where the movement is not depicted within the image frame.}
    \label{fig:1}
\end{figure}

Several challenges were introduced into the dataset to enhance its complexity. First, there were considerable variations in the time intervals between successive frames, with some intervals spanning up to 1.5 minutes, resulting in limited overlapping visual features. Secondly, the frame rate differed significantly across different trajectories, leading to noticeable discontinuities in the dataset. To streamline the dataset, the lower portion of each image, which predominantly depicted static car components, was excluded from consideration as it was deemed irrelevant for pose estimation tasks. Furthermore, the dataset exhibited diverse lighting conditions, necessitating that models developed by participants demonstrate robustness in handling such variations. Figure~\ref{fig:1} illustrates the challenges within the dataset. Firstly, on the left, we have a standard pair of images (1 sec apart) that are relatively easy to match. In the center is a typical pair (2 sec apart) that presents more difficulty in matching due to sharp turns, which occur infrequently in the sequence, making them particularly challenging. Finally, on the right is a pair of images (76 sec apart) that can hardly be matched.

\begin{figure}
    \centering
    \includegraphics[width=0.9\linewidth]{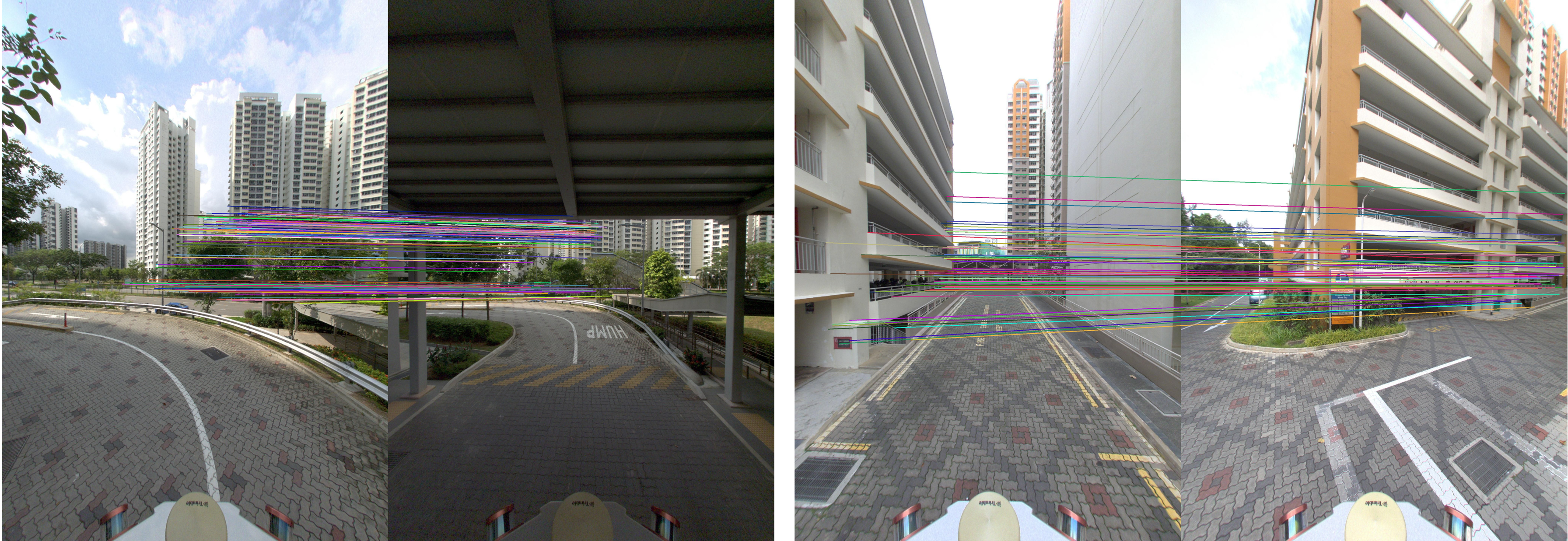}
    \caption{Illustration of challenging matches by LightGlu.}
    \label{fig:2}
\end{figure}

Despite being consecutive in the sequence, some images display characteristics such as time gaps in the dataset, where non-overlapping pairs may result in the method generating arbitrary relative poses, leading to subpar outcomes in straightforward sequential matching.  In Figure~\ref{fig:2}, we illustrate challenging matches addressed by LightGlue~\cite{lindenberger2023lightglue}, a deep neural network tailored to matching sparse local features between image pairs. In the left image, numerous matching points are obscured by the roof, while in the right image, a $\ang{90}$ turn of the car causes matching failure.

\section{Proposed Methods}

The AISG–SLA Visual Localization Challenge took place at IJCAI 2023 from May 26, 2023, to July 26, 2023, featuring a total prize pool of up to USD 40,000. With over 300 participants globally, the event saw the formation of more than 50 teams. In this section, we provide an overview of the winning teams' strategies and their approaches to addressing the pose estimation problem. Due to space limitations, readers are encouraged to contact the teams for further details~\footnote{\url{https://aisingapore.org/aisg-sla-visual-localisation-challenge-winners/}}.

\subsection{RoMa and DeDoDe Strategy}

GETINGARNA, the first-place winning team, scored 0.0273 in rotational error and 1.4205 in translational error, leverages their recent research on deep learning-based image matching, notably the RoMa~\cite{edstedt2023roma} and DeDoDe methods~\cite{edstedt2024dedode}. These techniques excel in reliably estimating relative poses for most consecutive image pairs within the challenge sequence.

\begin{figure}
    \centering
    \includegraphics[width=\linewidth]{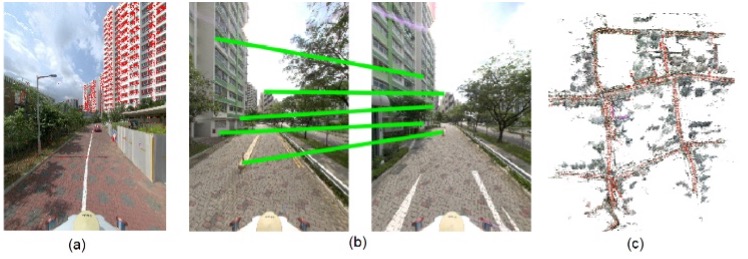}
    \caption{Method pipeline. 3(a) Extract DeDoDe keypoints in all images and fine non-sequential image pairs to match using i) DINO v2, ii) manual inspection. 3(b) Match keypoints in sequential and non-sequential image pairs using RoMa, filter with Graph-Cut RANSAC. 3(c) Structure from motion using COLMAP, each red dot is a position where a picture was taken.}
    \label{fig:3}
\end{figure}

To attain accurate estimates for the most challenging image pairs and address loop closures. The team devised a pipeline that integrates image retrieval with DINOv2~\cite{oquab2023dinov2}, and full structure-from-motion reconstruction with COLMAP~\cite{schonberger2016structure,schonberger2016pixelwise}. By employing image retrieval with DINOv2, they successfully matched specific non-consecutive image pairs, effectively addressing the challenge of significant time gaps in the image sequence. These improvements contributed to their solution outperforming all competitors.

\begin{figure}
    \centering
    \includegraphics[width=0.55\linewidth]{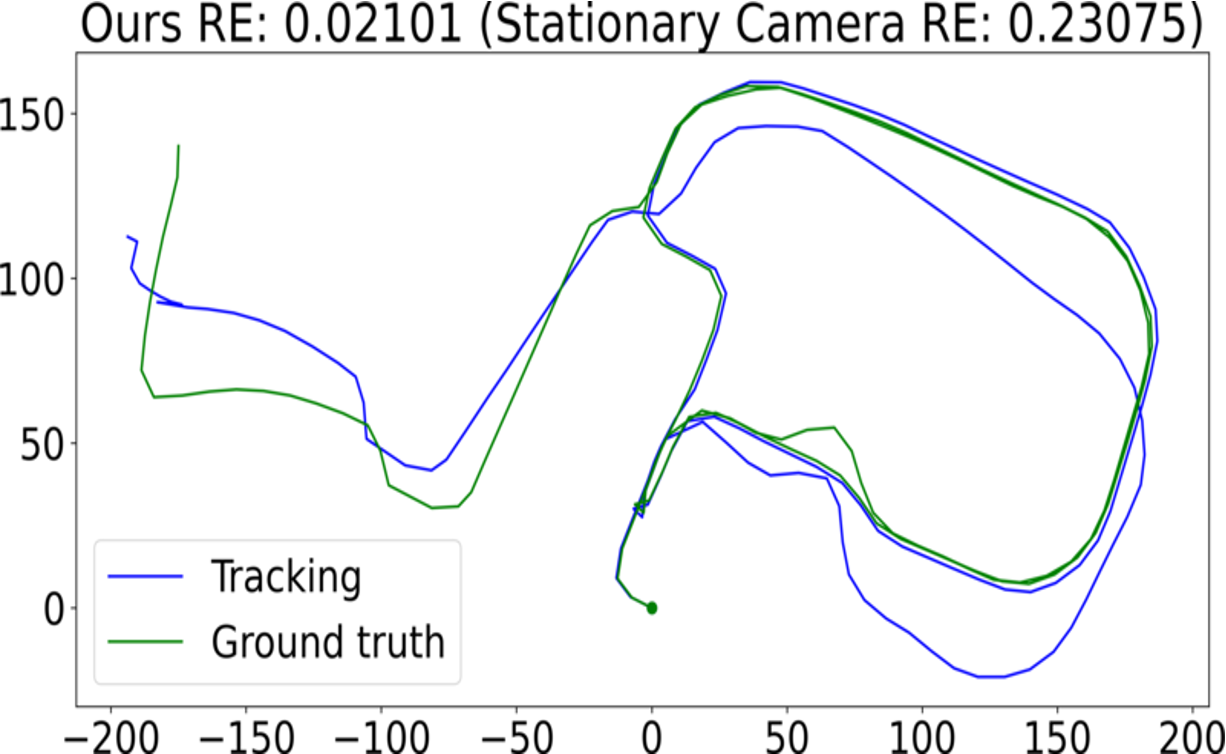}
    \caption{Visual representation of predicted sub-trajectories of the rotational error (RE).}
    \label{fig:4}
\end{figure}

KBRODT, securing 2nd place (Figure~\ref{fig:4}), adopted a similar approach, leveraging RoMa and the deep neural network model DeDoDe descriptor upon recognizing the limitations of ORB detectors~\cite{campos2021orb} and FLANN matchers~\cite{muja2009fast} for the task at hand. Their strategy revolves around utilizing publicly available pretrained weights for both RoMa and DeDoDe descriptor models. This strategic decision not only streamlined their implementation process but also eliminated the necessity for extensive re-training, ensuring stable and robust performance across various tasks. Notably, the team achieved a rotational error of 0.0382 and a translational error of 6.3374.

\subsection{CNN-based Strategy}

The third-place winning teams, Team SYLISH and Team SDRNR (tied), both adopted a CNN-based strategy in their architectural designs. SYLISH's solution integrated LightGlue and MobileNetV2~\cite{sandler2018mobilenetv2} as fundamental components, establishing a deep image retrieval network and deep feature matching for 2D-2D correspondence among similar images. They inferred relative motion between frames through essential matrix estimation within a RAN-SAC scheme and estimated translation scale factors using a deep-scale estimation network. The resulting poses underwent refinement via nonlinear least-squares optimization to address rotation graph problems.

SDRNR utilized basic CNN approaches like EfficientNet and ResNet. These models processed two RGB images from consecutive states, accompanied by precomputed depth maps and flow maps, predicting quaternions for rotation and corresponding translations. They advanced their model by integrating a visual matching system capable of identifying anchor point matches on RGB images, even amidst significant camera pose changes. Additionally, they implemented heuristics to address situations where no matches were found, particularly during $\ang{180}$ turns or anomalously long time deltas between states.

\section{Conclusion}

3D mapping research is crucial for smart city development, but obtaining 3D data is expensive. Monocular camera position estimation provides a cost-effective solution by determining camera pose from visual cues. The AISG–SLA Visual Localization Challenge (VLC) at IJCAI 2023 showed that state-of-the-art techniques from the research community can be effective in accurately extracting camera pose data from 2D images in the real world. Furthermore, the VLC dataset, available for research purposes, provides a valuable resource for further research in dynamic urban environments.

\section*{Acknowledgements}

AISG - This research/project is supported by the National Research Foundation, Singapore under its AI Singapore Programme.

Team Getingarna (Georg Bökman, Johan Edstedt) - This work was supported by the Wallenberg Artificial Intelligence, Autonomous Systems and Software Program (WASP), funded by the Knut and Alice Wallenberg Foundation and by the strategic research environment ELLIIT funded by the Swedish government.

Team kbrodt (Kirill Brodt) - We acknowledge Novosibirsk State University for providing the GPU server.

\bibliographystyle{named}
\bibliography{ijcai24}

\begin{thebibliography}{}

\bibitem[\protect\citeauthoryear{Barros \bgroup \em et al.\egroup }{2022}]{barros2022comprehensive}
Andr{\'e}a~Macario Barros, Maugan Michel, Yoann Moline, Gwenol{\'e} Corre, and Fr{\'e}d{\'e}rick Carrel.
\newblock A comprehensive survey of visual slam algorithms.
\newblock {\em Robotics}, 11:24, 2022.

\bibitem[\protect\citeauthoryear{Burri \bgroup \em et al.\egroup }{2016}]{burri2016euroc}
Michael Burri, Janosch Nikolic, Pascal Gohl, Thomas Schneider, Joern Rehder, Sammy Omari, Markus~W Achtelik, and Roland Siegwart.
\newblock The euroc micro aerial vehicle datasets.
\newblock {\em The International Journal of Robotics Research}, 35(10):1157--1163, 2016.

\bibitem[\protect\citeauthoryear{Campos \bgroup \em et al.\egroup }{2021}]{campos2021orb}
Carlos Campos, Richard Elvira, Juan J~G{\'o}mez Rodr{\'\i}guez, Jos{\'e}~MM Montiel, and Juan~D Tard{\'o}s.
\newblock Orb-slam3: An accurate open-source library for visual, visual--inertial, and multimap slam.
\newblock {\em IEEE Transactions on Robotics}, 37(6):1874--1890, 2021.

\bibitem[\protect\citeauthoryear{Coors \bgroup \em et al.\egroup }{2000}]{coors2000matching}
Volker Coors, Tassilo Huch, and Ursula Kretschmer.
\newblock Matching buildings: Pose estimation in an urban environment.
\newblock In {\em Proceedings IEEE and ACM International Symposium on Augmented Reality (ISAR 2000)}, pages 89--92. IEEE, 2000.

\bibitem[\protect\citeauthoryear{Edstedt \bgroup \em et al.\egroup }{2023}]{edstedt2023roma}
Johan Edstedt, Qiyu Sun, Georg B{\"o}kman, M{\aa}rten Wadenb{\"a}ck, and Michael Felsberg.
\newblock Roma: Revisiting robust losses for dense feature matching.
\newblock {\em arXiv preprint arXiv:2305.15404}, 2023.

\bibitem[\protect\citeauthoryear{Edstedt \bgroup \em et al.\egroup }{2024}]{edstedt2024dedode}
Johan Edstedt, Georg Bökman, Mårten Wadenbäck, and Michael Felsberg.
\newblock {DeDoDe: Detect, Don't Describe --- Describe, Don't Detect for Local Feature Matching}.
\newblock In {\em 2024 International Conference on 3D Vision (3DV)}. IEEE, 2024.

\bibitem[\protect\citeauthoryear{Engel \bgroup \em et al.\egroup }{2014}]{engel2014lsd}
Jakob Engel, Thomas Sch{\"o}ps, and Daniel Cremers.
\newblock Lsd-slam: Large-scale direct monocular slam.
\newblock In {\em European conference on computer vision}, pages 834--849. Springer, 2014.

\bibitem[\protect\citeauthoryear{Engel \bgroup \em et al.\egroup }{2017}]{engel2017direct}
Jakob Engel, Vladlen Koltun, and Daniel Cremers.
\newblock Direct sparse odometry.
\newblock {\em IEEE transactions on pattern analysis and machine intelligence}, 40(3):611--625, 2017.

\bibitem[\protect\citeauthoryear{Geiger \bgroup \em et al.\egroup }{2013}]{geiger2013vision}
Andreas Geiger, Philip Lenz, Christoph Stiller, and Raquel Urtasun.
\newblock Vision meets robotics: The kitti dataset.
\newblock {\em The International Journal of Robotics Research}, 32(11):1231--1237, 2013.

\bibitem[\protect\citeauthoryear{Gonz{\'a}lez-Sabbagh and Robles-Kelly}{2023}]{gonzalez2023survey}
Salma~P Gonz{\'a}lez-Sabbagh and Antonio Robles-Kelly.
\newblock A survey on underwater computer vision.
\newblock {\em ACM Computing Surveys}, 55(13s):1--39, 2023.

\bibitem[\protect\citeauthoryear{Lindenberger \bgroup \em et al.\egroup }{2023}]{lindenberger2023lightglue}
Philipp Lindenberger, Paul-Edouard Sarlin, and Marc Pollefeys.
\newblock Lightglue: Local feature matching at light speed.
\newblock In {\em Proceedings of the IEEE/CVF International Conference on Computer Vision}, pages 17627--17638, 2023.

\bibitem[\protect\citeauthoryear{Muja and Lowe}{2009}]{muja2009fast}
Marius Muja and David~G Lowe.
\newblock Fast approximate nearest neighbors with automatic algorithm configuration.
\newblock {\em VISAPP (1)}, 2(331-340):2, 2009.

\bibitem[\protect\citeauthoryear{Mur-Artal \bgroup \em et al.\egroup }{2015}]{mur2015orb}
Raul Mur-Artal, Jose Maria~Martinez Montiel, and Juan~D Tardos.
\newblock Orb-slam: a versatile and accurate monocular slam system.
\newblock {\em IEEE transactions on robotics}, 31(5):1147--1163, 2015.

\bibitem[\protect\citeauthoryear{Oquab \bgroup \em et al.\egroup }{2023}]{oquab2023dinov2}
Maxime Oquab, Timoth{\'e}e Darcet, Th{\'e}o Moutakanni, Huy Vo, Marc Szafraniec, Vasil Khalidov, Pierre Fernandez, Daniel Haziza, Francisco Massa, Alaaeldin El-Nouby, et~al.
\newblock Dinov2: Learning robust visual features without supervision.
\newblock {\em arXiv preprint arXiv:2304.07193}, 2023.

\bibitem[\protect\citeauthoryear{Rosinol \bgroup \em et al.\egroup }{2020}]{rosinol2020kimera}
Antoni Rosinol, Marcus Abate, Yun Chang, and Luca Carlone.
\newblock Kimera: an open-source library for real-time metric-semantic localization and mapping.
\newblock In {\em 2020 IEEE International Conference on Robotics and Automation (ICRA)}, pages 1689--1696. IEEE, 2020.

\bibitem[\protect\citeauthoryear{Sandler \bgroup \em et al.\egroup }{2018}]{sandler2018mobilenetv2}
Mark Sandler, Andrew Howard, Menglong Zhu, Andrey Zhmoginov, and Liang-Chieh Chen.
\newblock Mobilenetv2: Inverted residuals and linear bottlenecks.
\newblock In {\em Proceedings of the IEEE conference on computer vision and pattern recognition}, pages 4510--4520, 2018.

\bibitem[\protect\citeauthoryear{Sch{\"o}nberger and Frahm}{2016}]{schonberger2016structure}
Johannes~L Sch{\"o}nberger and Jan-Michael Frahm.
\newblock Structure-from-motion revisited.
\newblock In {\em Proceedings of the 2016 IEEE Conference on Computer Vision and Pattern Recognition (CVPR 2016)}, pages 4104--4113. IEEE, 2016.

\bibitem[\protect\citeauthoryear{Sch{\"o}nberger \bgroup \em et al.\egroup }{2016}]{schonberger2016pixelwise}
Johannes~L Sch{\"o}nberger, Enliang Zheng, Jan-Michael Frahm, and Marc Pollefeys.
\newblock Pixelwise view selection for unstructured multi-view stereo.
\newblock In {\em Computer Vision--ECCV 2016: 14th European Conference, Amsterdam, The Netherlands, October 11-14, 2016, Proceedings, Part III 14}, pages 501--518. Springer, 2016.

\bibitem[\protect\citeauthoryear{Wang \bgroup \em et al.\egroup }{2020}]{wang2020tartanair}
Wenshan Wang, Delong Zhu, Xiangwei Wang, Yaoyu Hu, Yuheng Qiu, Chen Wang, Yafei Hu, Ashish Kapoor, and Sebastian Scherer.
\newblock Tartanair: A dataset to push the limits of visual slam.
\newblock In {\em 2020 IEEE/RSJ International Conference on Intelligent Robots and Systems (IROS)}, pages 4909--4916. IEEE, 2020.

\bibitem[\protect\citeauthoryear{Zhou \bgroup \em et al.\egroup }{2018}]{zhou2018deeptam}
Huizhong Zhou, Benjamin Ummenhofer, and Thomas Brox.
\newblock Deeptam: Deep tracking and mapping.
\newblock In {\em Proceedings of the European conference on computer vision (ECCV)}, pages 822--838, 2018.

\bibitem[\protect\citeauthoryear{Zubizarreta \bgroup \em et al.\egroup }{2020}]{zubizarreta2020direct}
Jon Zubizarreta, Iker Aguinaga, and Jose Maria~Martinez Montiel.
\newblock Direct sparse mapping.
\newblock {\em IEEE Transactions on Robotics}, 36(4):1363--1370, 2020.

\end{thebibliography}

\end{document}